\documentclass[10pt,twocolumn,letterpaper]{article}

\usepackage{cvpr}
\usepackage{times}
\usepackage{epsfig}
\usepackage{graphicx}
\usepackage{amsmath}
\usepackage{amssymb}
\usepackage{booktabs}
\usepackage{tabularx}

\usepackage[pagebackref=true,breaklinks=true,letterpaper=true,colorlinks,bookmarks=false]{hyperref}

 \cvprfinalcopy 

\def\cvprPaperID{****} 

\ifcvprfinal\fi
\begin{document}

\title{Realistic Image Generation using Region-phrase Attention}

\author{Wanming Huang\\
University of Technology, Sydney\\
{\tt\small wanming.huang@student.uts.edu.au}
\and
Yida Xu\\
University of Technology, Sydney\\
{\tt\small yida.xu@uts.edu.au}
\and
Ian Oppermann\\
NSW Treasury\\
{\tt\small ianopper@outlook.com}
}
\maketitle

\begin{abstract}
   The Generative Adversarial Network (GAN) has recently been applied to generate synthetic images from text. Despite significant advances, most current state-of-the-art algorithms are regular-grid region based; when attention is used, it is mainly applied between individual regular-grid regions and a word. These approaches are sufficient to generate images that contain a single object in its foreground, such as a ``bird'' or ``flower''. However, natural languages often involve complex foreground objects and the background may also constitute a variable portion of the generated image. Therefore, the regular-grid based image attention weights may not necessarily concentrate on the intended foreground region(s), which in turn, results in an unnatural looking image. Additionally, individual words such as ``a", ``blue" and ``shirt" do not necessarily provide a full visual context unless they are applied together. For this reason, in our paper, we proposed a novel method in which we introduced an additional set of attentions between true-grid regions and word phrases. The true-grid region is derived using a set of auxiliary bounding boxes. These auxiliary bounding boxes serve as superior location indicators to where the alignment and attention should be drawn with the word phrases. Word phrases are derived from analysing Part-of-Speech (POS) results. We perform experiments on this novel network architecture using the Microsoft Common Objects in Context (MSCOCO) dataset and the model generates $256 \times 256$ conditioned on a short sentence description. Our proposed approach is capable of generating more realistic images compared with the current state-of-the-art algorithms.
\end{abstract}

\section{Introduction}

Generating images from text descriptions is a challenging problem that has attracted much interest in recent years. Algorithms based on the Generative Adversarial Network (GAN) \cite{NIPS2014_5423}, specifically Deep Convolutional GAN (DCGAN) \cite{DBLP:journals/corr/RadfordMC15}, have demonstrated promising results on various datasets. 

\cite{DBLP:journals/corr/RadfordMC15} synthesized images based on the RNN network encoded text descriptions and were able to generate $64 \times 64$ images on CUB, Oxford-102 and MSCOCO datasets. The RNN network was pre-trained using a deep convolutional and recurrent text encoder introduced by \cite{DBLP:journals/corr/ReedASL16}. The Generative Adversarial What-Where Network (GAWWN) \cite{NIPS2016_6111} was another GAN network that introduced extra information, such as bounding boxes and key points, and enabled the location of the main object in an image to be controllable \cite{NIPS2016_6111}. Following on from this work, \cite{zhang2016stackgan} proposed StackGAN which was able to generate photo-realistic $256 \times 256$ images for text descriptions through a 2-stage generation process. \cite{DBLP:journals/corr/abs-1711-10485} further extended StackGAN by incorporating an attention mechanism to perform multi-stage image generation. This method allowed attention to be paid to relevant words in generating different regions of images. 

These works have demonstrated a significant breakthrough in synthesizing images that contain a single object, such as the CUB \cite{WelinderEtal2010} and Oxford-102 \cite{Nilsback08} datasets in which each image contained a single specific type of flower or bird. However, synthesizing an image that models human poses or involves multi-object interactions usually lacks sufficient details, and can easily be distinguished from real images.

We believe that the in-depth connection between individual words and image sub-regions is not yet fully utilized and the model performance could be improved upon. For example, consider the sentence: \textit{A man swinging a baseball bat.} We would expect \textit{a man} and \textit{a baseball bat} and their interaction, \textit{swinging} are all to be captured in the generated image, while retaining some degree of freedom in the direction of the action, and/or the exact position for each object in the image.

In this paper, a few novel strategies have been proposed based upon the attention mechanism introduced by AttnGAN framework \cite{DBLP:journals/corr/abs-1711-10485}. First, we adjusted the Deep Attentional Multimodal Similarity Model (DAMSM) loss by utilizing true-grid features inside each bounding box in addition to phrases that consist of multiple consecutive words. This revised loss function encouraged the network to learn the in-depth relationship between a sentence and its generated image: in terms of both how the whole image reflected the prescribed sentence, as well as how specific regions of an image related to an individual word or phrase.

We also incorporated the bounding box and phrase information in defining attention weights between image regions and each word phrase. Therefore when generating pixels inside an object bounding box, the attention was paid to phrases such as ``a red shirt'' or ``a green apple'', instead of focusing on each individual word separately.

The rest of this paper is organised as follows. In section \ref{section:review}, we review the GAN network and several other literatures that we applied as the basis and inspiration of our work. In section \ref{section:architecture}, we introduce assumptions and the architecture of our model. The performances are compared and discussed in section \ref{section:discussion}. 

\section{Background and Related Work} \label{section:review}
In this section we review previous works on text embedding, GAN network structure, Region of Interest (RoI) Pooling and image-sentence alignment that we used as the basis for our work.

\subsection{Sentence Embedding}
Generating images from text requires each sentence to be encoded into a fixed length vector. Previous works such as StackGAN used text embedding generated by a pre-trained Convolutional Recurrent Neural Network \cite{DBLP:journals/corr/ReedASL16}. The RNN network has been widely applied in modelling natural languages for classification or translation purposes \cite{DBLP:journals/corr/SutskeverVL14}.

\subsection{Text-to-Image with GAN}
\label{t2i}
The Generative Adversarial Network involves a 2-player non-cooperative game by generator and discriminator. The generator produces samples from the random noise vector $z$, and the discriminator differentiates between true samples and fake samples. The value function of the game is as follows:
\begin{equation}
   \begin{aligned}
      \underset{G}{min} \underset{D}{max} V (D,G) &= \mathbb{E}_{x \sim p_{data}(x)} [log D(x)] \\
      &+ \mathbb{E}_{z \sim p_{z}(z)} [log (1 - D(G(z)))]
   \end{aligned}
\end{equation}

Deep Convolutional GAN (DCGAN) \cite{DBLP:journals/corr/RadfordMC15} utilized several layers of convolutional neural networks to encode and decode images, in addition to Batch Normalization \cite{DBLP:journals/corr/IoffeS15} to stabilize the GAN training.

On the basis of DCGAN, GAN-CLS \cite{DBLP:journals/corr/ReedAYLSL16} generated images based on the corresponding image caption $t$ in addition to the noise vector $z$. $z$ was sampled from a Gaussian distribution $z \in \mathbb{R}^{Z} \sim \mathcal{N}(0,1)$, and the text description $t$ was encoded with a pre-trained text encoder $\varphi$ to be $\varphi(t)$. $\varphi(t)$ was then concatenated with $z$ and processed through series of transposed convolution layers to generate the fake image. In the discriminator $D$, a series of convolution-batch normalization-leaky ReLU were applied to discriminate between true images and fake images.

GAWWN \cite{NIPS2016_6111} proposed an architecture for controllable text-to-image generation that adopted supplementary information such as bounding boxes or part locations of the main object in the image.

As previous works failed to generate images with higher-resolution than $128 \times 128$, StackGAN \cite{zhang2016stackgan} employed a 2-stage GAN network to generate photo-realistic $256 \times 256$ images from text descriptions. Its architecture consisted of 2 stages: Stage-I generated a low-resolution image (e.g. $64 \times 64$) based on texts and random noises, Stage-II generated a higher resolution image (e.g. $256 \times 256$) based on texts and the lower resolution images from Stage-I. 


AttnGAN \cite{DBLP:journals/corr/abs-1711-10485} was an extension of StackGAN, which used an attention mechanism in addition to an image-text matching score. It was able to generate images of better quality and achieve higher inception score. 

The generation process of AttnGAN was based on a multi-stage attention mechanism. At each stage, the generated image received information from attention weights. These attention weights were calculated between image features from the last stage and text features. The attention mechanism was also used to calculate a Deep Attentional Multimodal Similarity Model (DAMSM) loss, which encouraged the correct matching between sentence-image pairs. These details are discussed in section \ref{embedding_loss}.

\subsection{RoI Pooling}
The Region of Interest (RoI) pooling layer was first introduced in \cite{Girshick_2015_ICCV}. For an image region with spatial size $h \times w$, it was first divided into $H \times W$ grids of sub-windows. Each sub-window was then fed through a max-pooling layer, which derived a final pooling result with spatial size $H \times W$. The RoI pooling allowed each image region to be embedded into a fixed-length vector with no additional parameters and training involved.

\section{Architecture} \label{section:architecture}

\begin{figure*}
  \includegraphics[width=\textwidth]{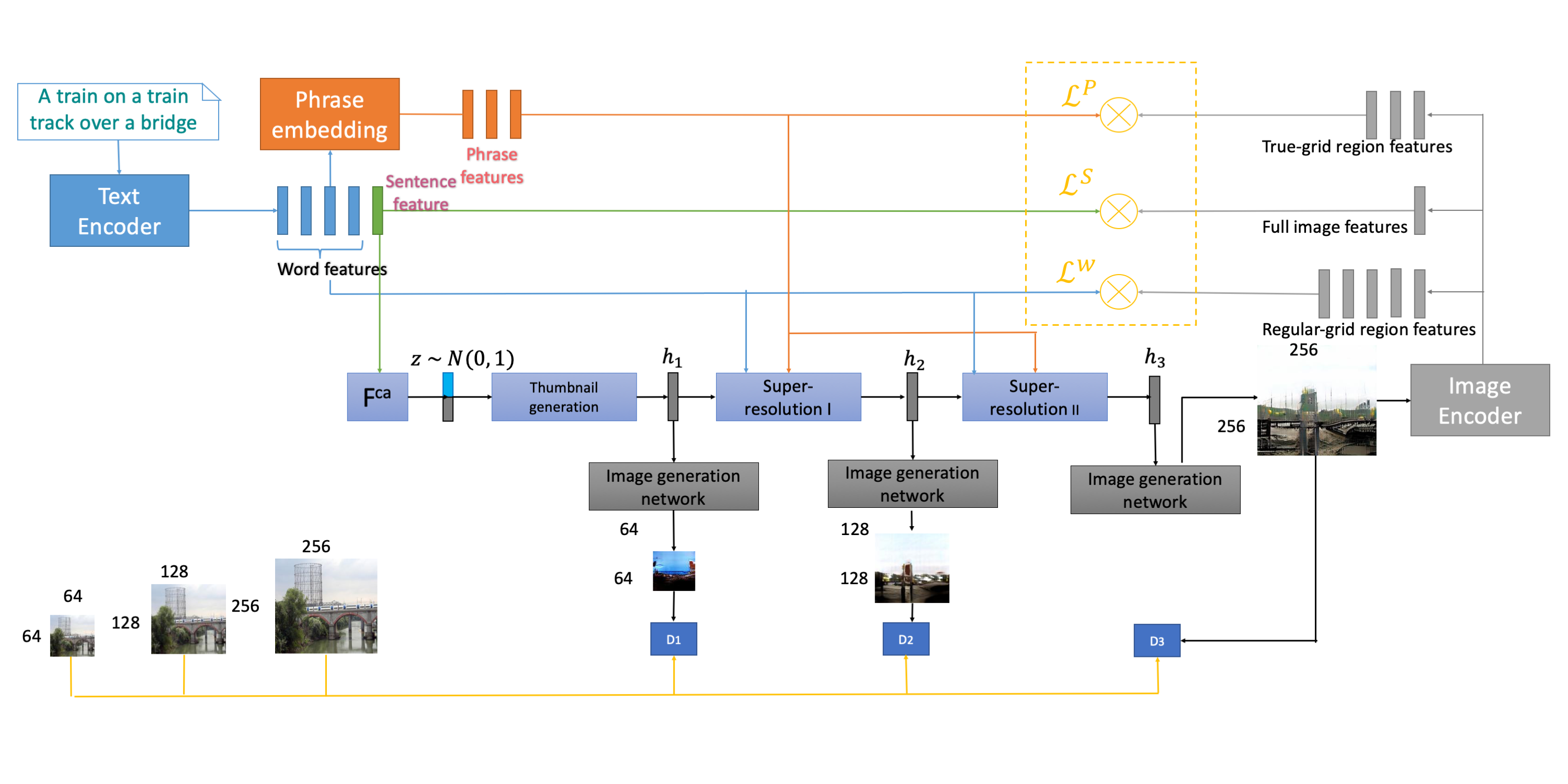}
  \caption{Network structure for sentence embedding and GAN network}
  \label{gan_graph}
\end{figure*}

We construct our network based on the latest architecture of AttnGAN \cite{DBLP:journals/corr/abs-1711-10485}. Inspired by the visual-semantic alignment \cite{Karpathy_2015_CVPR}, we also encourage image sub-region and word matching during training. As shown in figure \ref{gan_graph}, the proposed architecture consists of an end-to-end text encoder network and a GAN framework. The details are explained in the following sections.

\subsection{Text Encoder} \label{textencoder}

Current GAN based text to image generation networks typically extract a whole sentence representation and word representations using a bi-directional LSTM \cite{zhang2016stackgan,DBLP:journals/corr/ReedAYLSL16,DBLP:journals/corr/abs-1711-10485}. Our proposed algorithm takes advantage of previous methods while extracting additional phrase features. 

We define a phrase as a combination of an article, adjective and noun. Such information can be extracted from raw sentences via part-of-speech tagging (POS-tagging). For example, a sentence {\it 'A man is riding a surfboard on a wave'} is tagged as [('A', indefinite article), ('man',noun),('is', verb),('riding',verb),('a',indefinite article),('surfboard',noun), ('on', preposition), ('a', indefinite article), ('wave', noun)].  We then group the nearest article-adjective-noun words as a phrase, which yields {\it ''a man''}, {\it ''a surfboard''} and {\it ''a wave'}.

\begin{figure}[h!]
  \includegraphics[width=0.5\textwidth]{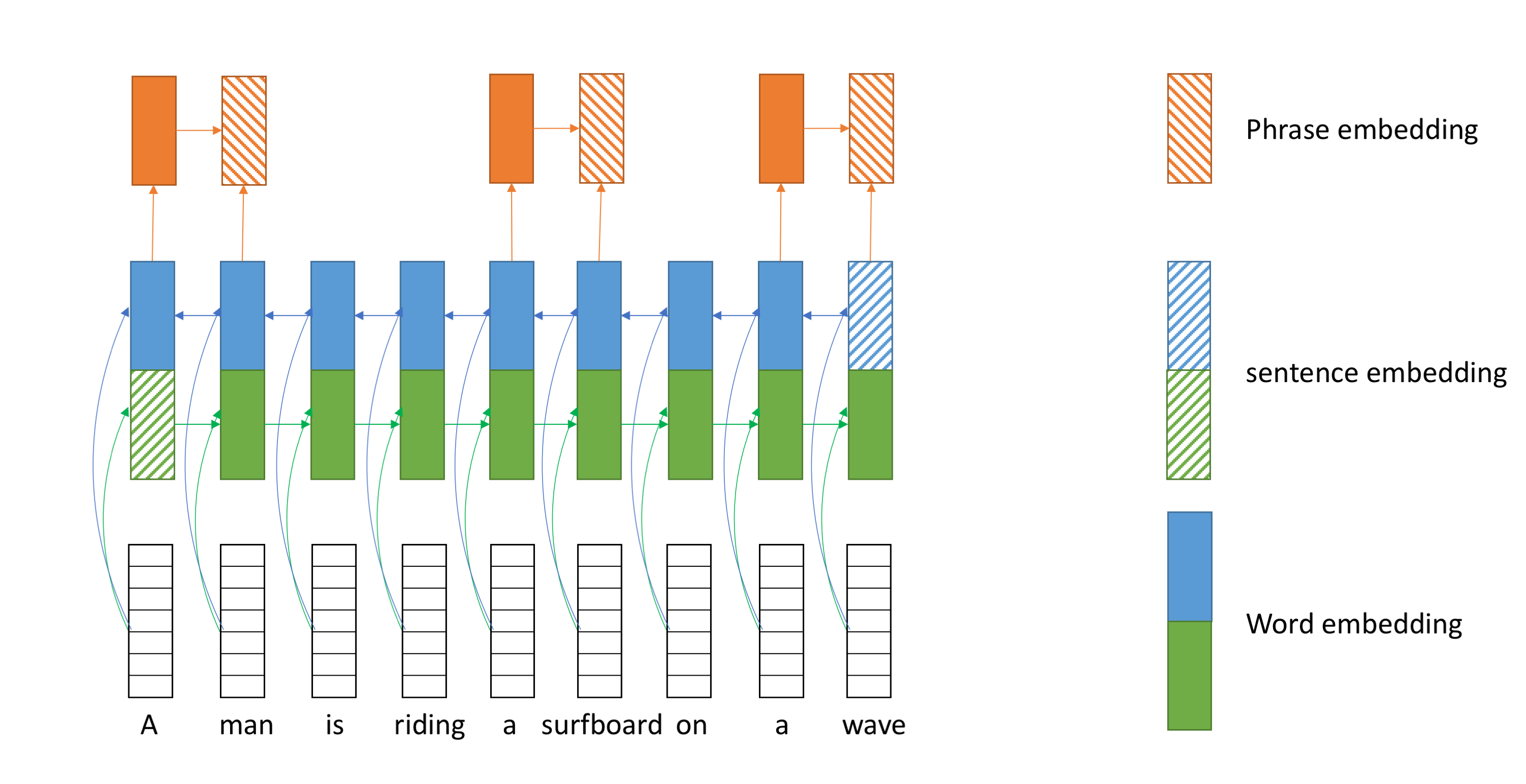}
  \caption{Text Embedding with 2 LSTM Networks}
  \label{text_embedding_graph}
\end{figure}

Training of the text encoder is assisted by the image encoder, which is explained in section \ref{imencoder} and section \ref{embedding_loss}. The full text encoder framework is shown in figure \ref{text_embedding_graph}. 

\subsubsection{Sentence Encoder}

Firstly, a bi-directional LSTM \cite{schuster1997bidirectional,graves2005bidirectional} is applied to each sentence to extract word and sentence representations. Given a sentence $\{w_1, \dots w_T \}$, the $t^{\text{th}}$ word representation $e_t$ is a concatenation of a forward $e^f_t$ and a backward hidden state $e^b_t$, i.e., $e_t \equiv [e^f_t \; e^b_t]$. The full sentence embedding is defined using the last hidden state $\bar{e}$. 

\subsubsection{Phrase Encoder}

On top of the extracted word representations $e$ and $e \equiv \{e_1, \dots e_T \}$, phrase representations are extracted by applying a second LSTM in the following way. Given the $t'^{th}$ phrase, a LSTM is applied over the sequence of words in the phrase. The last hidden state is used as its feature representation which we refer to as $p_{t'}$.

Our phrase-based embedding clearly has an advantage over the traditional word-based mechanism where each word has a seperate representation. For example, none of the {\it individual} words in the phrase {\it ``a green apple''} portrays an overall picture of the object; all three words work together to capture its visual meaning.

An alternative method to encode phrases is through averaging word embeddings. We compared the performance of the proposed text embedding against the original design from AttnGAN \cite{DBLP:journals/corr/abs-1711-10485} in addition to this simple word concatenation in section \ref{expdamsm}.

\subsubsection{Image Encoder}
\label{imencoder}

The image encoder itself comes from the pre-trained Inception-v3 network \cite{DBLP:journals/corr/SzegedyVISW15} and is not further fine-tuned. In our work, we apply the image encoder to extract three different image features from a single image: a {\it true-grid region feature}, a {\it regular-grid region feature } and a {\it full image feature}. 

\begin{figure}[h!]
  \includegraphics[width=0.5\textwidth]{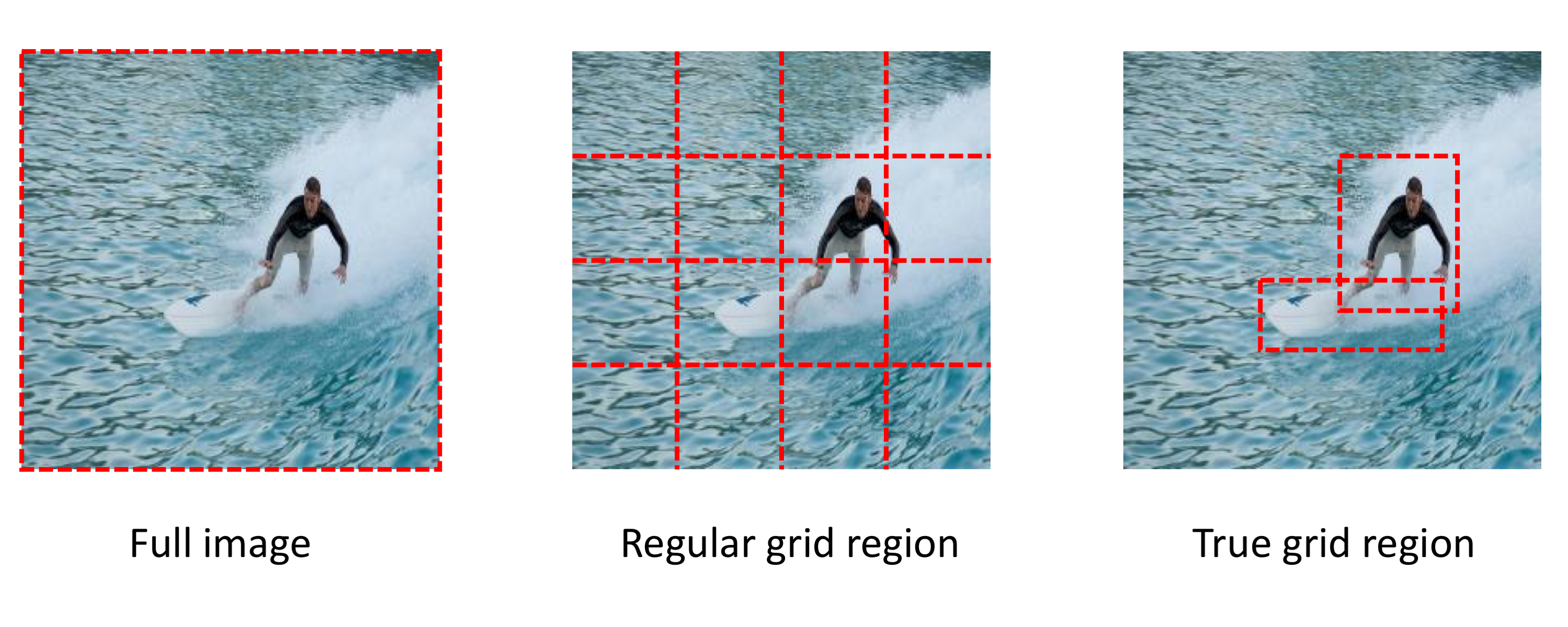}
  \caption{Examples of full image region, regular-grid region and true-grid region}
  \label{region_exp}
\end{figure}

Examples of each region are shown in figure \ref{region_exp}. A true-grid region is defined over an single object and thus the regions differ in sizes. Regular-grid regions have equal sizes and each of them can contain half or multiple objects.

Common to all features, each image first undergoes a pre-trained Inception-v3 model \cite{DBLP:journals/corr/SzegedyVISW15}. We use the ''$mixed\_6e$'' layer feature map as the designated layer for the regular-grid region. The full image feature is obtained from the last average pooling layer. In addition, both {\it regular-grid region feature} and {\it full image feature} are converted into vectors in the same semantic space using a Fully Connected (FC) layer. Therefore, the resulting features have the following dimensions: a regular-grid region feature $v \in R^{289 \times D}$ where $289 = 17 \times 17$ is the dimension for ''$mixed\_6e$'' layer feature map. The image feature is denoted as $\bar{v} \in R^{D}$.

To obtain a {\it true-grid region feature}, first we need the location and size of each region. In several open datasets, such as MSCOCO, the manually-labeled bounding boxes of object(s) within an image are readily available. In the case where the dataset does not provide such information, they can also be obtained from off-the-shelf image object detectors, such as RCNN \cite{Girshick_2014_CVPR}. This makes it possible to apply our algorithm to any image datasets with text annotations, including CUB and Oxford-102.

The ''$mixed\_6e$'' layer feature map and its bounding box information is fed through the Region of Interest (ROI) pooling to generate its true-grid region feature. The extracted feature after the ROI pooling is the same size despite the fact that each bounding box may differ in sizes. These features are fed through a convolution operation with a kernel of an equivalent size, resulting in a vector in a common semantic space as text features. We denote the true-grid region feature as $b \in R^{K \times D}$ where $K$ is the number of bounding boxes in each image. 

\subsubsection{Attention Based Embedding Loss} \label{embedding_loss}

Text embedding and the perceptron layer for image and region features are bootstrapped prior to training the GAN network. 

Following  \cite{DBLP:journals/corr/abs-1711-10485}, the training target is to minimize the negative log posterior probability for the correct image-sentence pair. i.e. for a batch of image-sentence pairs ${(S_{i}, I_{i})}_{i=1}^{M}$, the loss function is given as:

\begin{equation}
	   \mathcal{L} = - \overset{M}{\underset{i=1}{\sum}} (log P(S_{i} | I_{i}) + P(I_{i} | S_{i}))
\end{equation} \label{eqembeddingloss}


$P(S_{i} | I_{i})$ is the posterior probability for a sentence $S_{i}$ to be matched with an image $I_{i}$. 

\begin{equation}
	P(S_{i} | I_{i}) = \frac{exp(\gamma_{1} R(S_{i}, I_{i}))}{\sum^{M}_{q=1} exp(\gamma_{1}R(S_{q}, I_{i}))}
\end{equation} \label{eqpprob}

Here $R(S_{i}, I_{i})$ gives the similarity score between the sentence and the image and $\gamma_{1}$ is a manually defined smooth factor. The posterior probability $P(I_{i} | S_{i})$ for an image being matched to a sentence is defined in a similar way. 

The similarity score $R(S_{i}, I_{i})$ is defined from three perspectives. The first score $R_{1}(S_{i}, I_{i})$ uses the cosine similarity between a sentence representation $\bar{e}$ and a whole image feature $\bar{v}_{i}$. The second is to utilise an attention mechanism built between the regular-grid regions and the words:

\begin{equation}
   R_{2}(S_{i}, I_{i}) = log(\overset{T}{\underset{t}{\sum}} exp(\gamma_{2}R(c_{t}, e_{t})))^{\frac{1}{\gamma_{2}}}
\end{equation}

$\gamma_{2}$ is a second smooth factor. $R(c_{t}, e_{t})$ is the cosine similarity between a word embedding $e_{t}$ and a region-context vector $c_{t}$ which is calculated as a weighted sum over regular-grid image features:

\begin{equation} \label{attneqct}
  c_{t} = \overset{289}{\underset{j=0}{\sum}}\alpha_{j}v_{j}, \text{where} \hspace{0.3cm} \alpha_{j} = \frac{\gamma_{3}\bar{s}_{t,j}}{\sum_{k}^{289} exp(\gamma_{3}\bar{s}_{t,k})}
\end{equation} 

In equation \ref{attneqct}, $\alpha_{j}$ is the attention weight for the $j^{th}$ regular-grid towards the $t^{th}$ word. $\bar{s}_{t,j}$ is a normalised cosine similarity between the word and the region. 

The third way to define $R(S_{i}, I_{i})$ is through the attention mechanism between the true-grid regions and the phrases:

\begin{equation}
	R_{3}(S_{i}, I_{i}) = log(\overset{K}{\underset{t'}{\sum}} exp(\gamma_{2}R(c_{t'}, p_{t'})))^{\frac{1}{\gamma_{2}}}
\end{equation}

Here $c_{t'}$ is the region context vector which is computed from a weighted sum over true-grid region features.

\subsubsection{Text Encoder Loss Functions}

Bringing the three different $R$ values to equation \ref{eqembeddingloss} and equation \ref{eqpprob}, the three loss functions are noted as $\mathcal{L}^{S}$ and $\mathcal{L}^{W}$ and $\mathcal{L}^{P}$ as shown in graph \ref{gan_graph}. Later in section \ref{expdamsm}, we compare different combination of these loss functions, where LSTM-BASIC is the baseline model using $\mathcal{L}_{TEXT} = \mathcal{L}^{S} + \mathcal{L}^{W}$, LSTM utlizes all three loss functions as $\mathcal{L}_{TEXT} = \mathcal{L}^{S} + \mathcal{L}^{W} + \mathcal{L}^{P}$; LSTM-PHRASE uses $\mathcal{L}_{TEXT} = \mathcal{L}^{S} + \mathcal{L}^{P}$.

\subsection{Attentional Text to Image Generation}

Following the work by AttnGAN \cite{DBLP:journals/corr/abs-1711-10485}, our work constructs text to image generation as a multi-stage process. At each generation stage, images from small to large scales are generated from corresponding hidden representations. At the first stage, the thumbnail generation takes sentence embedding $\bar{e}$ as the input and generates images with the lowest resolution. At the following stages, images with higher resolution are generated through an attention structure. Details are explained below.

\subsubsection{Thumbnail Generation}

\begin{figure*}[h!]
  \includegraphics[width=\textwidth]{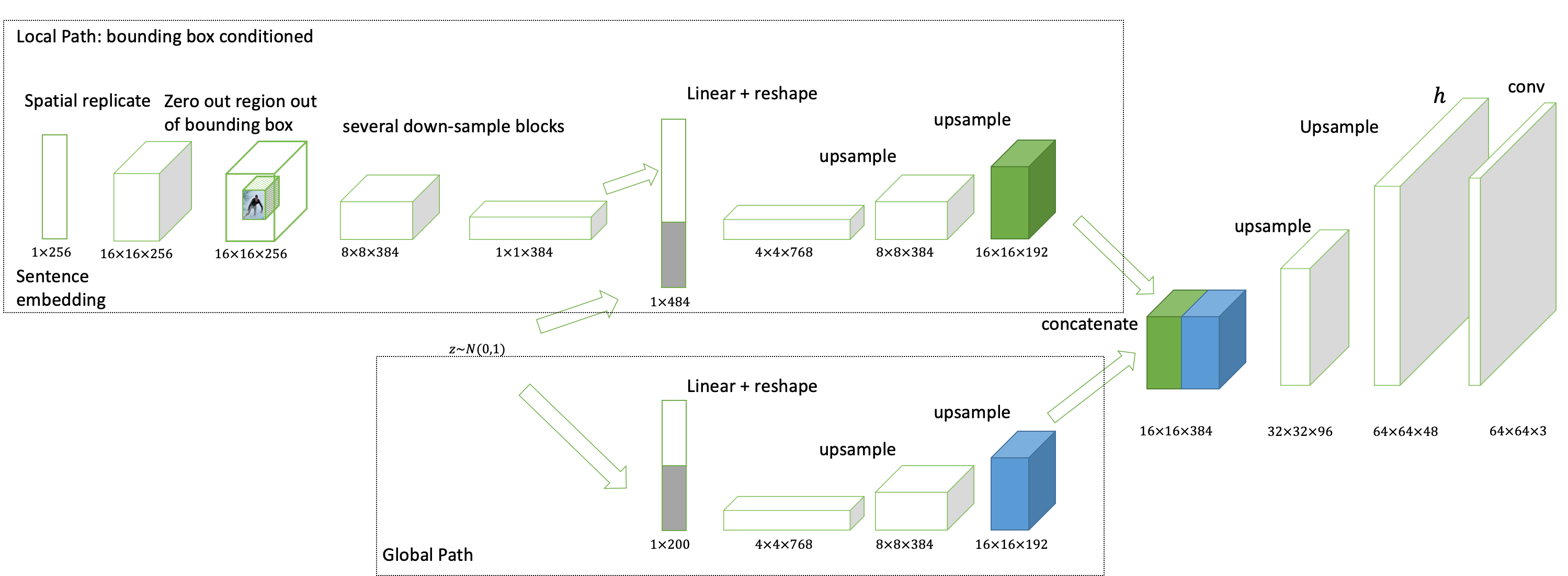}
  \caption{Thumbnail Generator}
  \label{imthumbnail}
\end{figure*}

The thumbnail generation is inspired from the vanilla sentence to image design by \cite{NIPS2016_6111}, which generates images conditioning on the sentence and additional information including bounding boxes and keypoints. The network structure is shown in figure \ref{imthumbnail}.

The generation process branches into two paths. The global path, which is not bounding box conditioned, takes the conditioning factor $F_{0}$, concatenates with the noise vector and fed through several upsampling block to a global feature tensor. The local path first spatially replicate the sentence embedding, and zeros out the region out of the bounding box. The masked text tensor is fed through upsampling blocks to a local feature tensor. Tensors from both paths are concatenated depth-wise and fed through another two upsampling blocks to derive $h_{1}$.

\subsubsection{Super-resolution I \& II}

Super-resolution enlarges the previously generated thumbnails through the attention mechanism. At stage $n$, a hidden representation $h_{n}$ is constructed from the last hidden state $h_{n-1}$. The hidden representation is later translated to an image with the \textit{image generation network} in section \ref{imggennet}. 

We incorporate two set of attentions in our framework, the first is between individual words and regular-grid regions, the second is between phrases and true-grid regions.

Given the word embeddings $e$ where $e \equiv {e_{1}, \dots e_{T}}$ for $T$ words in a sentence and phrase embeddings $p$ where $p \equiv {p_{1}, \dots p_{T'}}$ for $T'$ phrases in a sentence, $h_{n}$ is calculated as:

\begin{equation}
   h_{n} = F_{n}(h_{n-1}, F^{attn1}_{n}(e, h_{n-1}), F^{attn2}_{n}(p, h_{n-1}))
\end{equation}

Here, $F_{n}$ is a deep neural network that constructs the hidden representation $h_{n}$ from given inputs, $F^{attn1}$ and $F^{attn2}$ are the deep neural networks that construct the word-context matrix and  phrase-context matrix respectively.

The word-context matrix is constructed from word representations  $\{e_1, \dots e_T \}$ and regular-grid image region features from $h_{n-1}$. Word embeddings are first fed through a perceptron layer to be converted into the common semantic space as image features. The regular-grid region is defined here in a similar way to section \ref{imencoder}, except that the input feature map is not from the pre-trained Inception-v3.

Given $j^{th}$ regular-grid region feature $h_{n-1}^{j}$, a word-context vector $c_{j} = \overset{T}{\underset{t}{\sum}} \varphi_{j,t} e_{t}$ is defined as the weighted sum over word embeddings: 

\begin{equation} \label{eqattnmap}
   \varphi_{j,t} = \frac{exp({h_{n-1}^{j}}^{\top} e_{t})}{\sum_{\tau}^{T} exp({h_{n-1}^{j}}^{\top} e_{\tau})} 
\end{equation}

Here $\varphi_{j,t}$ is the attention weight between the $t^{th}$ word and the $j^{th}$ regular-grid region. Suppose there are $J$ regular-grids, the final word-context matrix is then defined as the union of the $c_{j}$ value for each regular-grid region, i.e., $F^{attn1}_{n}(e, h_{n-1}) = (c_{1}, \cdots , c_{J})$.

This phrase-context matrix is calculated in a similar way as in equation \ref{eqattnmap}, except that word embeddings are replaced with phrase features, and regular-grids are replaced with true-grid features. Here true-grid features are derived from $h_{n-1}$ by feeding it through the RoI pooling. 

The resulting matrix is of length $K$ where $K$ is the number of objects defined in the image. In order to apply such a matrix to the network, we let each pixel inside the bounding box carry the same phrase context vector while pixels outside of bounding box carry zeros. As for regions where multiple bounding boxes overlap, the phrase context vectors are averaged. The resulting phrase-context matrix is of the same shape as the previously defined word-context matrix. These two context matrices are again averaged to generate the final hidden representation.

\subsubsection{Image Generation Network: Hidden Representation to Images} \label{imggennet}

As shown in figure \ref{gan_graph}, the previous thumbnail generation and super-resolution stages do not produce images directly, they instead produce hidden representations that are fed through an additional convolution operation with kernel size and the depth dimension $3$ to generate images.

%

\subsubsection{Discriminators} \label{discriminators}


In general, we use three types of discriminators. The first evaluates a given image as being real or fake, the second evaluates a pair of image and sentence, and the third evaluates a group of image, sentence and bounding boxes. In addtion, we incorporate the logic of matching-aware discriminator from \cite{DBLP:journals/corr/ReedAYLSL16}, where the latter two discriminators are fed through real, fake and false samples. 

Therefore the value function for the generator and the discriminator at each stage is given as:

\begin{equation}
   \begin{aligned}
      \scriptscriptstyle \underset{G_{i}}{min} \underset{D_{i}}{max} V (D_{i},G_{i})  &\scriptscriptstyle = \scriptscriptstyle \mathbb{E}_{x_{i} \sim p_{data}(x_{i})} [log D(x_{i})] + \scriptscriptstyle \mathbb{E}_{z \sim p_{z}(z)} [log (1 - D_{i}(G_{i}(z)))]\\
      &\scriptscriptstyle + \scriptscriptstyle \mathbb{E}_{x_{i} \sim p_{data}(x_{i})} [log D(x_{i}, \bar{e})] \scriptscriptstyle + \mathbb{E}_{z \sim p_{z}(z)} [log (1 - D_{i}(G_{i}(z), \bar{e}))]\\
      &\scriptscriptstyle + \scriptscriptstyle \mathbb{E}_{x_{i} \sim p_{data}(x_{i})} [log D(x_{i}, \bar{e}, box)] \\&\scriptscriptstyle + \mathbb{E}_{z \sim p_{z}(z)} [log (1 - D_{i}(G_{i}(z), \bar{e}, box))]
   \end{aligned}
\end{equation}

\subsection{Bounding Box Prediction} \label{bboxprediction}

As the image generation relies on bounding box information, a third bounding box prediction network is trained based on text embeddings from section \ref{textencoder}. We define two prediction tasks in the network, the first is to predict coordinates for bounding boxes. The second is to predict the number of bounding boxes. We structure both prediction as a regression problem from the sentence embedding. Therefore, given a sentence embedding, it is first fed through 2 seperate multi-layer neural networks, in which is the final layer of both networks is a mean squared error of the predicted value and the real value.

We adopt several processing step on the data in the following manner. First, the coordinates of bounding boxes is normalized to the proportion of the full length, so that the maximum value is $1$ regardless of the size of the bounding box or the image. Second,  given a predicted number of bounding boxes, the coordinates for the bounding boxes that out-numbers the predicted value are considered as ''invalid'', and are thus excluded in computing the loss. In addition, we define words or phrases such as ``left'', ``right'', ``on top of'' as position related words. In section \ref{experimentbboxprediction}, we report the performance of whether or not the training is only performed on sentences that contain position related words.

%
%
%



\section{Experiments} \label{section:discussion}

Below we demonstrate the performance of the revised text encoder and the proposed GAN network.

The dataset we used is the MSCOCO dataset, which includes various images that involve natural scenes and complex object interactions.  It contains 82,783 images for training and 40,504 for validation. Each image has 5 corresponding captions. Bounding boxes are provided for objects over 80 categories. 

Experiment results demonstrated below are performed on $30,000$ random samples from the validation set.  Two metrics, inception scores and r-precision are utilized to perform the evaluation.

\subsection{Metrics: Inception score and R-precision}
In terms of the metrics, as it is difficult to measure the performance of image generation in a quantitative way, the \textit{inception score} was proposed by \cite{NIPS2016_6125}.
\begin{equation}
   I = exp(\mathbb{E}_{x}D_{KL}(p(y|x)||p(y)))
\end{equation}
where $x$ is a generated image and $y$ is the label predicted by the Inception model \cite{DBLP:journals/corr/SzegedyVISW15}.

However, the inception score only measures the quality and diversity of images generated. It does not evaluate how accurate an image can reflect the description of a sentence. Therefore, previously another metric called R-precision is proposed. 

In AttnGAN, authors define R-precision to be the top $r$ relevant text descriptions out of $R$ retrieved texts for an image. The candidate sentences are one relevant and $99$ randomly selected sentences. The final R-precision is an average over $30,000$ generated images. However, we observed through our experiments that as the number of randomly selected unmatched sentences increases, R-precision values decrease. This value is also affected by how similar candidate sentences are. Therefore, we also report a second R-precision value with one ground truth ($R = 1$) and all the rest of the sentences ($29,999$ false samples) to be the mismatching samples. We denote the first value as R-precision(100), and the second value as R-precision(30K).

\subsection{The Revised Text Encoder} \label{expdamsm}

Below we demonstrate the performance of the revised text encoder by comparing R-precision scores of the validation set in the text encoder training. We compare the performance of three encoders which are different combinations of encoder loss values introduced in section \ref{embedding_loss}. 

Table \ref{textembrprecision} shows that in training the text encoder, introducing phrase and real-grid regions encourages finding the correct matching between real image-sentence pairs. Using phrase and true-grid regions to construct the loss and emit word and regular-grid regions further improves the result.

\begin{table}[h!]
    \centering

    \begin{tabular}{c|c}
    Experiment & \scriptsize R-precision (100)\\
    \hline
    \scriptsize LSTM& \scriptsize $0.7299 \pm 0.0450$ \\
    \scriptsize LST-BASIC & \scriptsize $0.7239 \pm 0.0426$ \\
    \scriptsize LSTM-PHRASE & \scriptsize $0.7306 \pm 0.0405$
    \end{tabular}

  \caption{R-precision(100) over training iterations for the text encoder. LSTM-BASIC is the basic bi-LSTM in the previous work. LSTM and LSTM-PHRASE comes from the proposed method.}
\label{textembrprecision}
\end{table}

\subsection{Bounding Box Prediction} \label{experimentbboxprediction}

In the phase of validation and training, the coordinate of each bounding box and number of bounding boxes are predicted by seperate networks, as explained in section \ref{bboxprediction}. 

In this section, we compare the performances of multiple alternatives of both predictions. The first is between applying 4 layers of neural nerworks on both tasks versus 1 layer. The second is whether use all sentences in the prediction tasks or use only those sentences with position related words. The definition of position related words is given in section \ref{bboxprediction}. Comparisons are made in terms of loss values on the validation set over training iterations.

\begin{figure}[h!]
  \centering
    \includegraphics[width=0.23\textwidth]{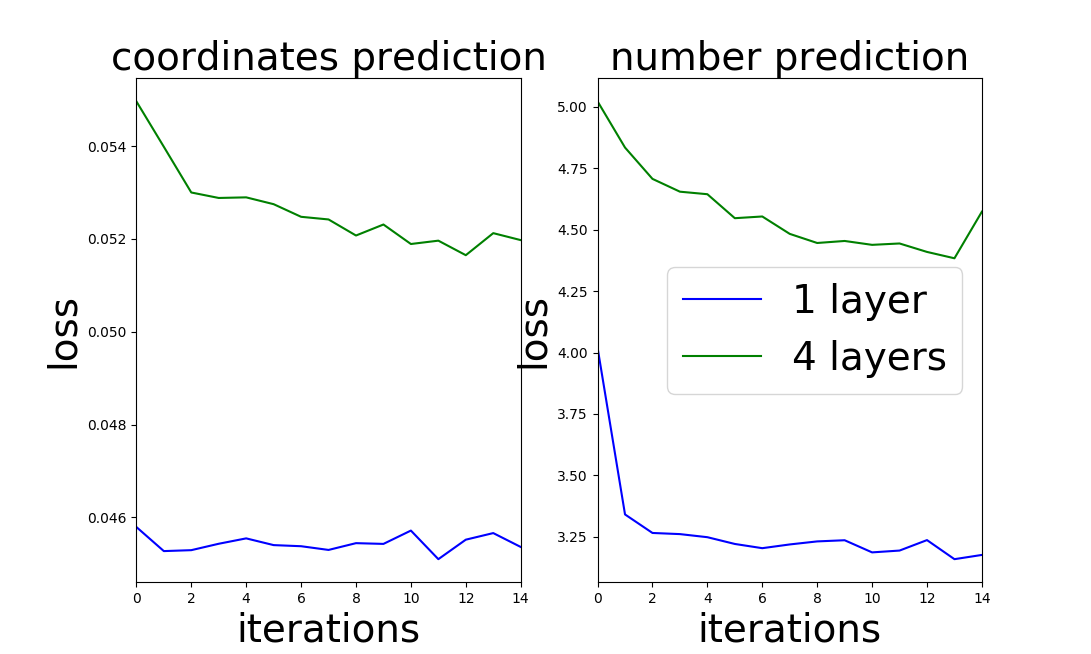}
    \includegraphics[width=0.23\textwidth]{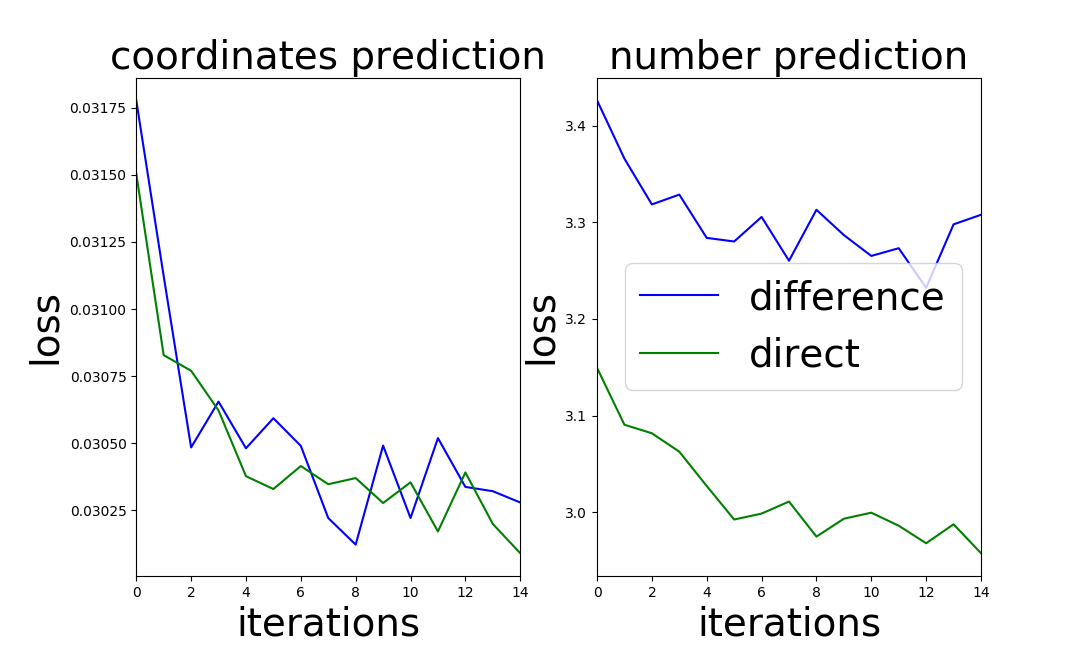}
    \caption{Comparison of multiple bounding box predictors. For both predictors, we report coordinates prediction loss and number prediction loss on the validation set over training iterations.} 
    \label{figbboxpred}
\end{figure}

From figure \ref{figbboxpred}, applying 4 layers on both predictions results in $0.0065$ and $1.2250$ higher loss than 1 layer, training only sentences with position related words improves the coordinates prediction by $0.0006$, but it results in a much higher loss in the number prediction task. Thus we can conclude that the best practice is applying 1 layer neural network to both predictions, and number of nouns should not be brought into the number prediction task.

\subsection{The GAN Network}


Table \ref{tabscore} reports the R-precision and Inception score achieved by the AttnGAN and the proposed method. The proposed method achieves  $1 \%$ higher R-precision(100) and $2 \%$ higher R-precision(30K). This shows that the proposed method is able to generate images that match more closely with the content described in the sentence.


\begin{table}[h!]
\resizebox{0.48\textwidth}{!}{%
\begin{tabular}{@{}llll@{}}
\toprule
Method   & R-precision(100) & R-precision(30K) & Inception Score(30K) \\ \midrule
AttnGAN  & $85.47 \pm 3.69 \% $ & $6.7 \%$ & $25.89 \pm 0.47 \%$  \\
Proposed & $86.44 \pm 3.38 \% $ & $9.5 \%$ & $23.74 \pm 0.36 \% $  \\ \bottomrule
\end{tabular}%
}
\caption{R-precision and Inception score between AttnGAN and the proposed method. }
\label{tabscore}
\end{table}

Table \ref{tabscore} also shows that the proposed method achieves a very close final Inception score to AttnGAN. Below we show three aspects with real samples that the generation result surpass previous methods.

Firstly, as shown in figure \ref{imggenmulti}, it is able to generate images that match closer with a given sentence, which is also proved by the higher R-precision rate. For example, when the generation is based on sentences that describe the number of objects explicitly, such as ``three sheep'' or ``adult and two children'', the proposed method is able to generate the correct number of objects in most cases. In addition, the proposed method is less likely to only focus on certain keywords and omit other important information. For example, in the case where ``stop sign'' and ``go light'' are both mentioned, the generated image displays both objects.

\begin{table}[]
\centering
\begin{tabular}{*{3}{m{0.13\textwidth}}}
\hline
Caption&AttnGAN&Proposed\\
\hline
A picture of a stop and go light with a stop sign next to it &\begin{center}\includegraphics[width=0.13\textwidth]{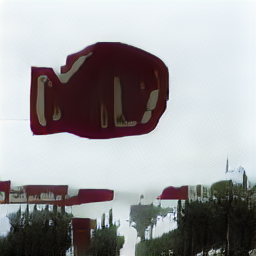} \end{center}&\begin{center}\includegraphics[width=0.13\textwidth]{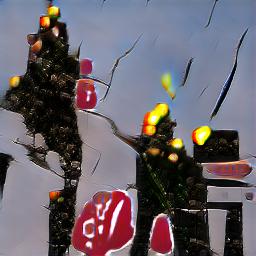} \end{center}\\
\hline
Three sheep in the process of running on grass outside &\begin{center}\includegraphics[width=0.13\textwidth]{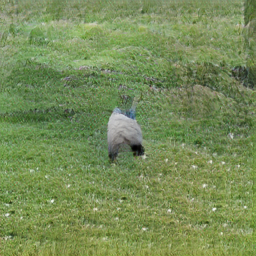} \end{center}&\begin{center}\includegraphics[width=0.13\textwidth]{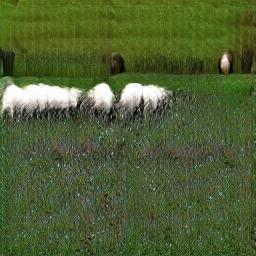} \end{center}\\
\hline
Adult and two children on the beach flying a kite &\begin{center}\includegraphics[width=0.13\textwidth]{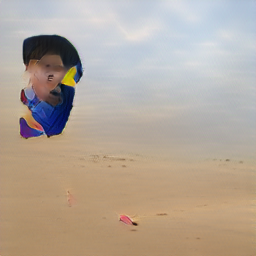} \end{center}&\begin{center}\includegraphics[width=0.13\textwidth]{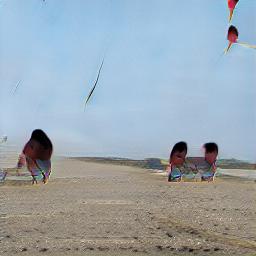} \end{center}\\
\hline
\end{tabular}
\caption{Example of images generated from a sentence that describes multiple objects.}
\label{imggenmulti}
\end{table}

Secondly, it is less likely for the proposed method to generate very similar images with different sentences, as opposed to AttnGAN. As shown in figure \ref{simimggen}, in the cases where ``surf'' or ``surfing'' related words are mentioned, AttnGAN generates very similar if not almost the same images. This issue also exists in cases of food related sentences. 

\begin{table}[]
\centering
\begin{tabular}{*{3}{m{0.13\textwidth}}}
\hline
Caption&AttnGAN&Proposed\\
\hline
A man riding a wave on a surfboard in the ocean &\begin{center}\includegraphics[width=0.13\textwidth]{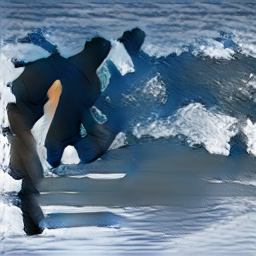} \end{center}&\begin{center}\includegraphics[width=0.13\textwidth]{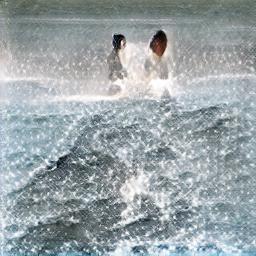} \end{center}\\
\hline
Person on surfboard laying on board going over a small wave &\begin{center}\includegraphics[width=0.13\textwidth]{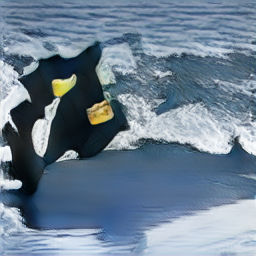} \end{center}&\begin{center}\includegraphics[width=0.13\textwidth]{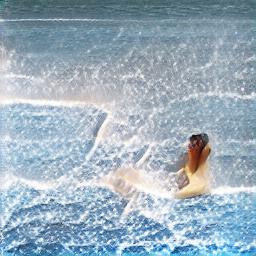} \end{center}\\
\hline
\end{tabular}
\caption{Examples of images generated from similar sentences.}
\label{simimggen}
\end{table}


Thirdly, table \ref{shapeimggen} shows that the proposed method performs better at displaying the correct shape of objects. 

\begin{table}[]
\centering
\begin{tabular}{*{3}{m{0.13\textwidth}}}
\hline
Caption&AttnGAN&Proposed\\
\hline
Large brown cow standing in field with small cow &\begin{center}\includegraphics[width=0.13\textwidth]{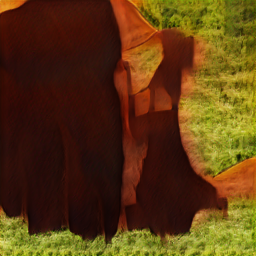} \end{center}&\begin{center}\includegraphics[width=0.13\textwidth]{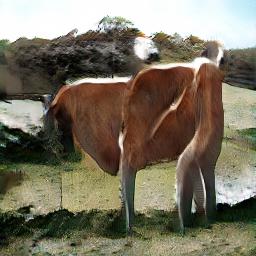} \end{center}\\
\hline
\end{tabular}
\caption{Example of images generated from a sentence where multiple words describe a certain core object.}
\label{shapeimggen}
\end{table}

Below we give another interesting example where our proposed method is actually able to generate back and white images with better quality.

\begin{table}[]
\centering
\begin{tabular}{*{3}{m{0.13\textwidth}}}
\hline
Caption&AttnGAN&Proposed\\
\hline
Black and white photo of a predestrian at a suburban crosswalk &\begin{center}\includegraphics[width=0.13\textwidth]{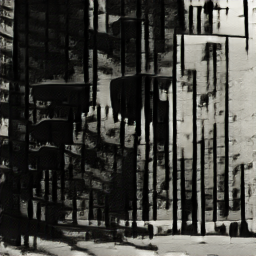} \end{center}&\begin{center}\includegraphics[width=0.13\textwidth]{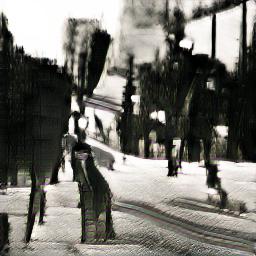} \end{center}\\
\hline
\end{tabular}
\caption{Example of images generated from a sentence where multiple words describe a certain core object.}
\label{shapeimggen}
\end{table}

\section{Summary}

Our work provides improvements on the state-of-the-art attention based GAN network for text to image generation. Our main contribution comes in two folds. Firstly, we propose a new design of text encoder which extracts additional phrase embeddings. Secondly, we incorporate a new set of attention which is between true-grid regions and phrases into the GAN network. Through the experimentation on MSCOCO dataset, our approach is capable of generating more realistic and accurate images.
{\small
\bibliographystyle{ieee}
\bibliography{gan}
}

\end{document}